\documentclass[runningheads]{llncs}
\usepackage{graphicx}
\usepackage{subfigure}

\usepackage{soul}
\usepackage{url}
\usepackage[utf8]{inputenc}

\usepackage{verbatim}
\usepackage{booktabs}
\usepackage{setspace}
\usepackage{color}
\usepackage{amsmath}
\usepackage{amssymb}
\usepackage{algorithm}
\usepackage{algorithmic}
\usepackage[backref]{hyperref}

\begin{document}
\title{Challenge Closed-book Science Exam: A Meta-learning Based Question Answering System}
\titlerunning{The MetaQA System for Closed-book Science Exam}
%
\author{Xinyue Zheng\inst{1} \and
Peng Wang\inst{1}\thanks{Corresponding author} \and
Qigang Wang\inst{1} \and Zhongchao Shi\inst{1}}
%
%
\institute{AI Lab, Lenovo Research, Beijing, 100089, China \\
\email{\{zhengxy10, wangpeng31, wangqg1, shizc2\}@lenovo.com}\\
}
\maketitle              

\begin{abstract}

Prior work in standardized science exams requires support from large text corpus, such as targeted science corpus from Wikipedia or SimpleWikipedia. However, retrieving knowledge from the large corpus is time-consuming and questions embedded in complex semantic representation may interfere with retrieval. Inspired by the dual process theory in cognitive science, we propose a MetaQA framework, where system 1 is an intuitive meta-classifier and system 2 is a reasoning module. Specifically, our method based on meta-learning method and large language model BERT, which can efficiently solve science problems by learning from related example questions without relying on external knowledge bases. We evaluate our method on AI2 Reasoning Challenge (ARC), and the experimental results show that meta-classifier yields considerable classification performance on emerging question types. The information provided by meta-classifier significantly improves the accuracy of reasoning module from $46.6\%$ to $64.2\%$, which has a competitive advantage over retrieval-based QA methods.

\keywords{Standardized science exam \and Meta-learning \and Question answering.}
\end{abstract}

\section{Introduction}
Standardized science exam is a challenging AI task for question answering \cite{clark2015elementary}, as it requires rich scientific and commonsense knowledge (see Table 1). Many researches \cite{clark2013study,jansen2017framing} solve these comprehensive science questions by retrieving from a large corpus of science-related text \cite{jansen2018worldtree}, which provide detailed supporting knowledge for the QA system \cite{musa2018answering,pan2019improving}. However, some questions are usually asked in a quite indirect way, requiring examiners to dig out the exact expected evidence of the facts. If the information of the correct viewpoint cannot be specifically extracted by the question representation, it may lead to incorrect information retrieval. Our work challenges closed-book science exams in which solvers do not rely on large amounts of supported text.

\begin{table}[!h]
\renewcommand\arraystretch{1.1} 
\centering  
\caption{Questions from ARC science exam, illustrating the need for rich scientific knowledge and logical reasoning ability. Bold font indicates the correct answer.}\label{tab1}
\begin{tabular}{|p{0.9\columnwidth}|} 
\hline
1. Which rapid changes are caused by heat from inside Earth? 
(A) landslides \textbf{(B) volcanoes} (C) avalanches (D) floods \\

2. What is one way to change water from a liquid to a solid? \textbf{(A) decrease the temperature} (B) increase the temperature (C) decrease the mass (D) increase the mass \\

3. How many basic units of information in a DNA molecule are required to encode a single amino acid? (A) 1 (B) 2 \textbf{(C)} 3 (D) 4 \\

\hline
\end{tabular}
\end{table}

How do human learn science knowledge? When solving science problems, the first step we take to find answers is to understand what the question is about \cite{hovy2001toward}, so we will make a simple mapping between the questions and the knowledge points we have learned, then use a few related examples to help inference. In the process of learning new knowledge day after day, we gradually master the skills of integrating and summarizing knowledge, which will in turn promote our ability to learn new knowledge faster. Dual process theory \cite{Dual-processing,empirical} in cognitive science suggests that human brain need two systems of thought. Intuitive system (System 1) is mainly responsible for fast, unconscious and habitual cognition; logic analysis system (System 2) is a conscious system with logic, planning, and reasoning. Inspired by the dual process theory, our work aims to build a human-like learning system to complete science exams in a more reasonable way. As shown in Figure 1, we propose a meta-learning based question answering framework (MetaQA), which consists of two systems. System 1 is a meta-learning module, which extracts meta-features from learning tasks to quickly classify new knowledge. System 2 adopts BERT, a large pre-trained language model with complex attention mechanisms, to conducts the reasoning procedure.

\begin{figure*}[!h]
\centering
\subfigure[Meta-learning dataset of one-shot classification.]{
\begin{minipage}[t]{0.4\linewidth}
\centering
\includegraphics[width = 1\linewidth]{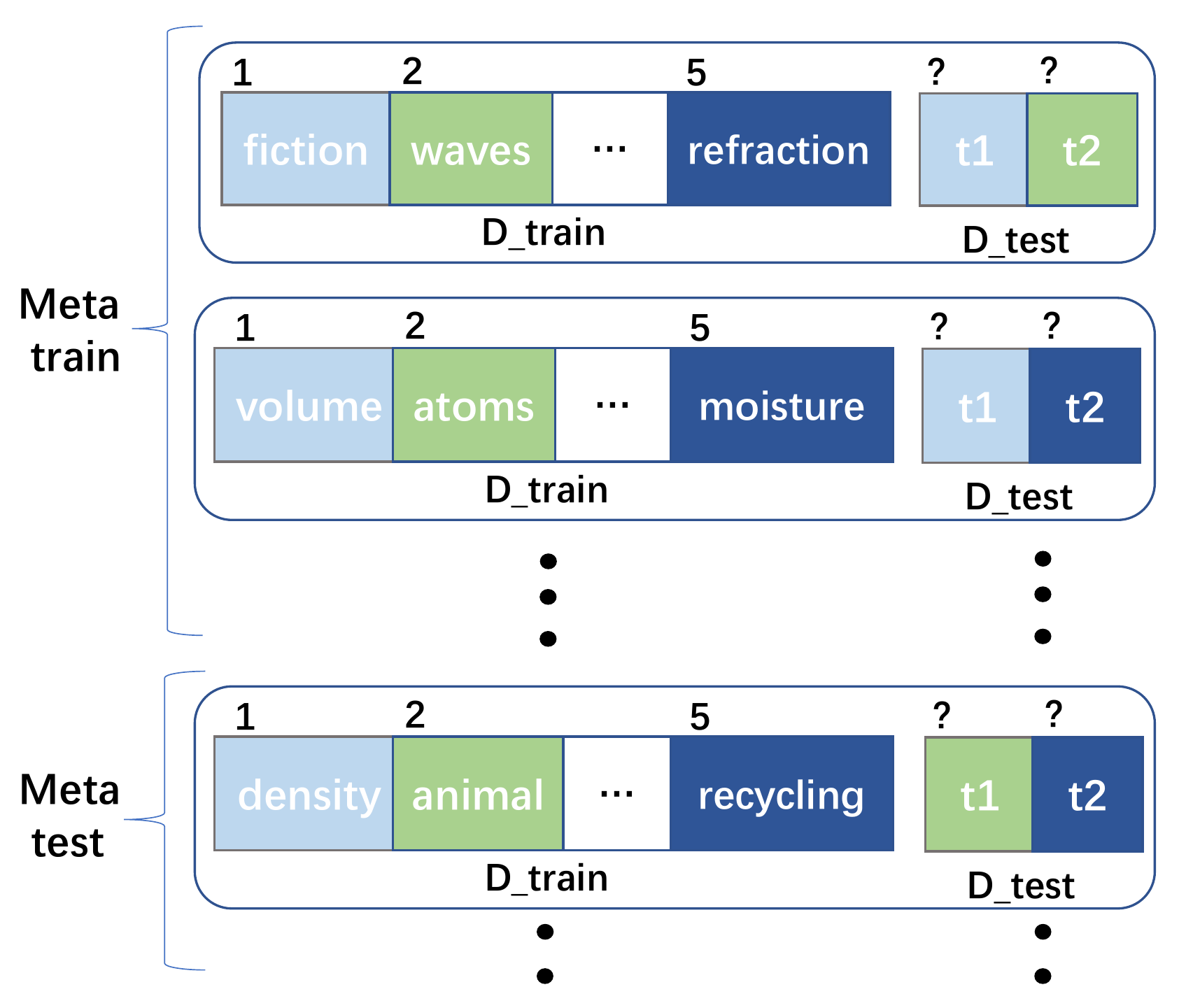}
\end{minipage}
}%
\hspace{.3in}
\subfigure[An overview of MetaQA implementation.]{
\begin{minipage}[t]{0.4\linewidth}
\centering
\includegraphics[width =  1\linewidth]{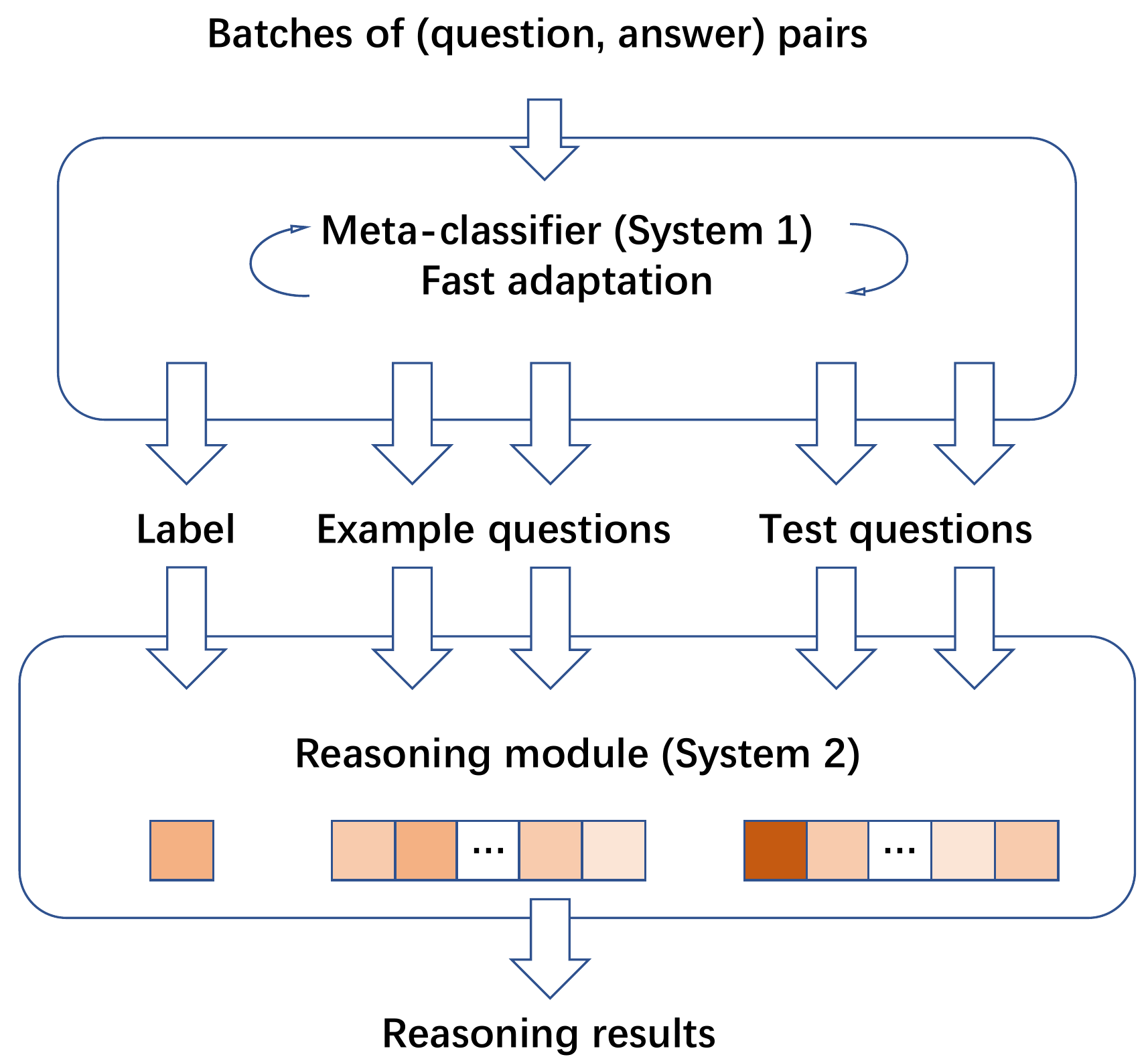}
\end{minipage}
}%
\hspace{.15in}
\centering
\caption{a): An example of meta-learning dataset. The meta-train set contains multiple tasks, and each task is a few-shot knowledge points classification problem. The key goal of meta-learning is to quickly classify new categories that have not been seen before. b): Overview of MetaQA system. The input of system 1 is the batches of different tasks in meta-learning dataset, and each task is intuitively classified through fast adaptation. System 2 uses classification information (label, example questions) given by system 1 to reason the test questions.}
\end{figure*}

At the first stage, we use model-agnostic meta-learning method (MAML) \cite{finn2017model} to train a meta-classifier which is able to capture the meta-information that contains similar features among different tasks, so that the meta-classifier can quickly adapt to new tasks with few samples. In this work, we regard questions with the same knowledge points $k$ as a task. At the second stage, the BERT model learns to reason testing questions with the assistance of question labels and example questions (examine the same knowledge points) given by the meta-classifier. Experiments show that the proposed method achieves impressive performance on the closed-book exam with the help of the few-shot classification information.

This work makes the following contributions:
\begin{itemize}
\item We are the first to consider closed-book science exam, and propose a MetaQA system to solve this challenging task according to human cognition. 

\item The MetaQA system does not rely on large corpus, which is applicable for practical situations when building a targeted knowledge base requires significant human workload and time costs.

\item Experimental results show that the proposed method greatly improves the QA accuracy by +17.6\%, and exceeds the performance of retrieval-based QA method.

\end{itemize}

\section{Related Work}

\subsection{Standardized Science Exams}

Various methods have been developed to tackle the science exams. Previous work \cite{khashabi2018question,clark2016my} was dominated by statistical, inference methods and information extraction. With the advent of large-scale language modeling \cite{devlin2018bert,liu2019roberta}, recent researchers who take advantage of this powerful word embedding model have been shown to improve performance. Qiu et al.\cite{ran2019numnet} propose NumNet+ to solve mathematical reasoning problems on DROP \cite{dua2019drop}. They use the number as graph nodes, then create directed edges based on the magnitude relationship of the numbers, and combined with RoBERTa \cite{liu2019roberta} to support complex mathematical reasoning functions. 

Many prior methods focus on the retrieval and analysis of problems, falling into the framework of knowledge base embedding and question answering from knowledge base. Clark et al.\cite{clark2019f} improve question answering with external knowledge accumulated from Wikipedia or targeted corpus. However, building a comprehensive corpus for science exams is a huge workload and complex semantic representation of questions may cause interference to the retrieval process. We propose a natural solution similar to people learning concepts, who cannot record all the background knowledge, but will classify the learned knowledge points and internalize the knowledge of related example questions as a tool for answering.

\subsection{Meta-learning}

Due to the huge cost involved in labeling questions, question classification datasets tend to be small \cite{roberts2014automatically,godea2018annotating}, which may create methodological challenges to taxonomy. Meta-learning seeks for the ability of learning to learn, by training through a variety of similar tasks and generalizing to new tasks with a small amount of data. Meta-learning approaches fall into two main categories: (1) Memory-based methods \cite{munkhdalai2017meta,santoro2016one}, which try to learn from past experience by adding external memory block on neural network. Santoro et al.\cite{santoro2016one} make use of Neural Turing Machine \cite{graves2014neural}, who is able to quickly encode new information, to combine information storage with retrieval functions to achieve rapid knowledge transfer and application. Simple Neural Attentive Learner (SNAIL) \cite{mishra2017simple} adopts temporal convolution to collect experience, and uses soft attention to pinpoint specific pieces of details for few-shot learning. (2) Meta-learner based methods \cite{nichol2018reptile,finn2017model} learn highly adaptive initialized parameters by repeatedly trained on a series of tasks, so that only a small number of gradients updates are required for fast learning a new task. In the fields of image classification, natural language processing \cite{gu2018meta,munkhdalai2017meta}, etc., MAML has proven to be a very effective meta-learning approach.

\section{Meta-classifier System}
\subsection{Few-shot Question Classification}

The ARC dataset contains 7,787 genuine grad-school level, 4-choice multiple choice science questions from 12 U.S. states over the past decade. The most interesting and challenging aspects of ARC exams is the multifaceted nature, with different questions examining different types of knowledge. Xu et al. \cite{xu2019multi} classifies science questions in ARC dataset according to their knowledge points and proposes the ARC classification dataset. Their work expands the taxonomy from 9 coarse-grained \emph{(e.g. life, forces, earth science, etc.)} to 406 fine-grained categories \emph{(e.g. migration, friction, Atmosphere, Lithosphere, etc.)} across 6 levels of granularity.

\begin{table}[!h]
\renewcommand\arraystretch{1.2} 
\renewcommand\tabcolsep{5.0pt}
\centering  
\caption{Data statistics of the ARC few-shot question classification dataset.}
\begin{tabular}{lccccc}  
\hline

Measure & L1 & L2 & L3 & L4 \\ 
\hline
\emph{Meta Train} \\
\hspace*{0.4cm}Categories   & $40$ &$92$  & $126$ &$150$ \\
\hspace*{0.4cm}QA Pairs   & $4,241$ &$4,461$  & $4,085$ &$3,705$ \\

\emph{Meta Test} \\
\hspace*{0.4cm}Categories  &  $33$ &$77$ & $104$  & $124$ \\
\hspace*{0.4cm}QA Pairs   & $3,496$ &$3,062$ & $3,287$  & $3,557$ \\

\hline
Total Categories  &$73$ &$169$ & $230$ & $274$ \\
Total QA pairs   & $7,737$ &$ 7,532 $ & $ 7,372$ & $ 7,262 $ \\

\hline
\end{tabular} 
\end{table}

\textbf{Few-shot QC Dataset:} We extract 4 levels \emph{(L2, L3, L4, and L6)} in ARC classification hierarchy provided by Xu et al., and reconstruct the few-shot question classification dataset according to the meta-learning settings. Firstly, We remove question categories with too few instances to do 5-shot learning (less than 6 samples). For each level, Meta-training set is created by randomly sampling around half classes from ARC dataset, and the remaining classes make up a meta-test set. Table 2 presents the data statistics of the ARC few-shot question classification dataset.

\textbf{Few-shot QC Model:}
Recent question answering work benefits from large-scale language models such as ELMo \cite{peters2018deep}, BERT \cite{devlin2018bert}, and RoBERTa \cite{liu2019roberta}. These models are pre-trained by predicting missing words or next sentence from huge amounts of text, and are robust enough to be fine-tuned to new NLP tasks, such as question classification and question answering. We make use of RoBERTa-base, a 12-layer language model with bidirectional encoder representations from transformers, as meta-classifier model.

For each question, we use question text $q_i$ and its correct answer option $a_i$ as RoBERTa input:

\hspace*{4.0cm} \emph{[CLS] {$q_{i}$} {$a_{i}$} [SEP].}

Where [SEP] is the token for sentence segmentation, and the [CLS] is a special token for output. The [CLS] output token for each question will be fed through a softmax layer, trained by cross-entropy loss against its correct label.

\subsection{Model Agnostic Meta-learning Method}

\textbf{Problem Definition:} For K-shot, N-way learning problem, meta-classifier learns a series of tasks from only K training sample per task (K $\leq$ 5). Define the $\mathcal{Q}$ as input question space, and $\mathcal{L}$ as label space. The meta-classifier with parameters $\theta$ can be presented as $f_\theta$: $\mathcal{Q}$ $\rightarrow$ $\mathcal{L}$, and solves the following few-shot classification tasks: 

\begin{equation}
\tau=\{\underbrace{(q_1, l_1),...,(q_{NK}, l_{NK})}_{\text{supporting samples * N tasks}},\underbrace{(q_t, l_t)}_{\text{query samples}},f_\theta\}
\end{equation}

For each task, specific supporting samples and query samples are formed to train a task-specific classification model $\tilde{f_\theta}$ by minimizing the loss function $L$($\tilde{f_\theta}$($q_t$),$l_t$).

\begin{figure}[h!]
\centering
\includegraphics[scale=0.45]{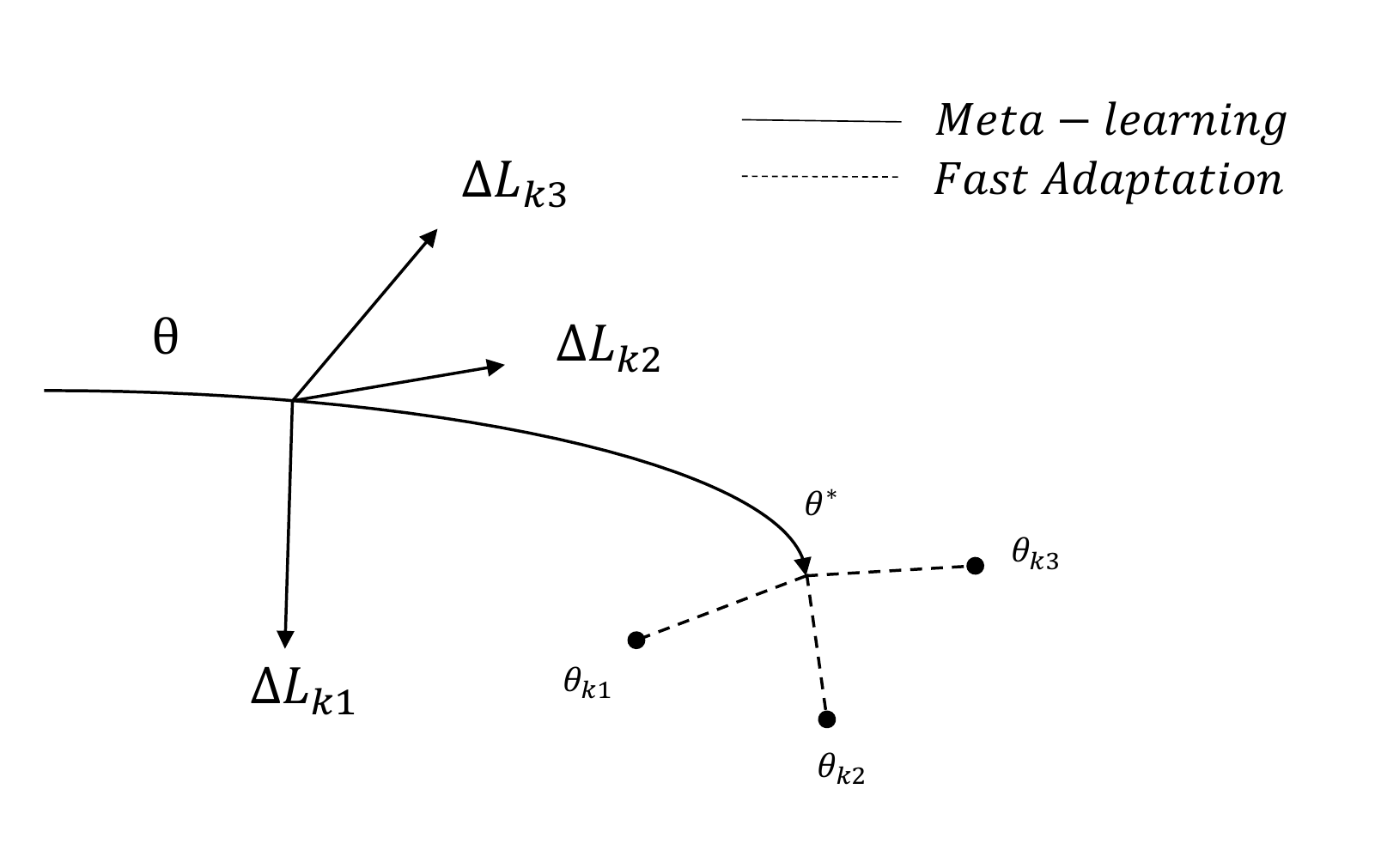}
\caption{Diagram of MAML algorithm, which trained through a series of tasks and set parameters at the optimal start point to adapt new tasks.}
\end{figure}

\textbf{Mete-learning Method:} We adopt MAML method to solve few-shot question classification. The MAML algorithm optimizes meta-learner at task level rather than data points. The meta-learner firstly learns well-initialized parameters $\theta$ in a distribution of specific tasks, and then meta-optimizes across each task's test samples by using updated parameters. The procedure can be presented in the following formula:

\begin{equation}
\min_{\theta}\sum_{\tau}{L(D_{\tau}^{'},\theta_{\tau}^{'})}=\sum_{\tau}{L(D_{\tau}^{'}, T(D_{\tau},\theta)).}
\end{equation}

Where $D_{\tau}$ and $D_{\tau}^{'}$ denote the support set and query set on task $\tau$ respectively, the $T(D_{\tau},\theta)$ is the training procedure acting on $D_{\tau}$, and the $L_{\tau}$ is computed on updated parameters $\theta'$ with query samples $D_{\tau}^{'}$. When applied to few-shot question classification, question queries with the same knowledge points $k$ are considered as task $\tau_k$. We make use of the amount of question classes to capture meta information, which includes common characteristics among different tasks $\tau_k$, so that the meta-information can be rapidly transferred to previously unseen classes. As shown in Figure 2, meta-classifier can be fast adapted to new models with parameters $\theta_{k1}$, $\theta_{k2}$ and $\theta_{k3}$ for the few-shot questions with knowledge points $k_1$, $k_2$ and $k_3$.

\begin{algorithm}[!h]  
    \caption{Meta-learning for few-shot question classification}  
    \begin{algorithmic}[1]  
        \STATE Randomly initialize $\theta$ for meta-classifier 
        \REPEAT
        \STATE Sample batch of knowledge points $K_{b} \sim p(K)$
        \FOR {each knowledge point $k$ in $K_{b}$}
        \STATE Sample support data $D$ and query data $D^{'}$ for task $\tau_k$
        \STATE Compute adapted parameters with $j$ gradients step: 
        \STATE $\theta_k^{'} = SGD(L_{\tau_k}, \theta, j)$ 
        \ENDFOR 
        \STATE Update $\theta \leftarrow \theta + \alpha\frac{1}{j}\sum\nolimits_{k=1}^K(\theta_k^{'} - \theta)$
        \UNTIL {Convergence}
    \end{algorithmic}  
\end{algorithm}

Algorithm $1$ details the process of optimizing initial parameters for meta-learning. In few-shot question classification setting, the model is trained to learn questions with batches of  knowledge points $K_{b} \sim p(K)$ sampled from support data $D$, and update the parameters to provide a good starting point for test samples from query data $D^{'}$. Because MAML method is model-agnostic, we can easily embed MAML into RoBERTa training procedure.

\section{Reasoning System}

In this section, we also choose RoBERTa as reasoning model, because its powerful attention mechanism can extract key semantic information to complete inference tasks. In order to apply RoBERTa more effectively, we firstly finetune model on RACE \cite{lai2017race} training set (87866 questions), a challenging set of English reading comprehension tasks in Chinese middle and high school exams. Through a single run using the same hyper-parameters in RoBERTa paper appendix, RoBERTa model gets scores of 75.9/75.1 on the validation set and test set. 

Coupling meta-classifier and reasoning system is a quite complicated process, we could either construct a great number of different solvers oriented to specific question types \cite{minsky1988society}, or make the meta-classifier produce incorporate information which can be directly used by reasoning model \cite{qiu1993concept}. Here, we adopt the latter method -- incorporating question classification information through query expansion.

We evaluate several different information expanding methods, including giving questions labels, using example questions, or combining both example questions and question labels as auxiliary information. Specifically, for a given pair, which includes test question ($q_{ti}$), example questions with the correct answer($q_{ei}$, $a_{ei}$), and question label ($l_i$), related information will be concatenated into the beginning of the question. There are three ways to achieve information expansion:

Only label information:

\hspace*{4cm} \emph{[CLS] {$l_i$} {$q_{ti}$} [SEP].}

Only example question:

\hspace*{4cm} \emph{[CLS] {$q_{ei}$} {$a_{ei}$} {$q_{ti}$} [SEP].}

Label and example question:

\hspace*{4cm} \emph{[CLS] {$l_i$} {$q_{ei}$} {$a_{ei}$} {$q_{ti}$} [SEP].}

\begin{table}[!h]
\renewcommand\arraystretch{1.1} 
\centering  
\caption{An example of query expansion for question classification labels.}
\begin{tabular}{p{0.99\columnwidth}}
\hline
Original Question Text\\
When air near the ground is warmed by sunlight, which of the following occurs?\\

\\
Expanded Text (Label information)\\
\textbf{Thermal Energy} When air near the ground is warmed by sunlight, which of the following occurs?\\
\\

Expanded Text (Example question information)\\
\textbf{Which is most responsible for the uneven heating of the air in the atmosphere? Convection} When air near the ground is warmed by sunlight, which of the following occurs?\\

\hline
\end{tabular}
\end{table}

Table 3 is an example of this process. Note that for labels under fine-grained classification, such as "LIFE\_REPROD\_COMPETITION" (Living Things -$\textgreater$ Reproduction -$\textgreater$ Competition), we only inform the reasoning model of the last level type (Competition).

The training process of question answering is similar to classification, where the [CLS] output token for each answer option is projected to a single logit, fed through a softmax layer, and trained through cross-entropy loss to get the correct answer.

\section{Experiments}

Table 2 in section 4 introduced the statistics of few-shot question classification dataset. For each classification level, we specify a certain category for training data, and remaining for test data. Taking L4 as an example, the meta-train set contains 150 categories with 3,705 training samples and the meta-test set consists of 124 categories with 3,557 test questions, and there is no overlap between training and testing categories.

\subsection{Few-shot Question Classification}

Few-shot question classification considers solving $K$-shot, $N$-way problem. We take $1$-shot, $5$-way classification as an example. For each task $\tau_i$, we firstly sample $25$ examples --- $1$(question) x $5$ (classes) to build a support set; then use MAML to optimize meta-classifier parameters on each task; and finally test our model on the query set which consists of test samples for each class. Because the question samples for each category is limited, few-shot question classification is a challenging task.

Table 4 shows that the MAML based question classification method achieves impressive performance in few-shot learning. We observe that model trained on L4 has better classification accuracy than L1. After detailed data analysis, we draw the following two inferences:

\emph{1) According to the experimental setting of meta-learning, certain tasks are randomly selected for each training time. Thus, a larger number of tasks tends to guarantee a higher generalization ability of the meta-learner. For L4 with the most tasks, it can generate a meta-classifier that is easier to quickly adapt to emerging categories.}

\emph{2) Although the L4 has most categories, the questions in the same fine-grained category have a high degree of similarity. Therefore, even if the model just fits the surface features (without generalization), it may achieve high accuracy.}

\begin{table*}[tbp]
\renewcommand\arraystretch{1.2} 
\renewcommand\tabcolsep{8pt}
\centering  
\caption{Empirical evaluation results (\%) of 5-shot 5-way question classification on the ARC science exam, broken down by classification granularity (L1 to L4).}
\begin{tabular}{llccccr}  
\hline
Scenarios & Method & L1 & L2 & L3 & L4 \\ 

\hline

$1$-shot $5$-way &Transfer learning & $45.8$ &$58.4$  & $61.5$ &$62.0$ \\

 &Meta-classifier & $61.8$ &$64.9$ & $69.8$ & $69.9$ \\

\hline
$5$-shot $5$-way &Transfer learning & $68.7$ &$78.1$  & $79.0$ &$80.7$ \\

 &Meta-classifier & $81.5$ &$82.6$ & $84.3$ & $85.5$ \\

\hline

\emph{Number of Categories} & & \emph{73} & \emph{169} & \emph{230} &\emph{274} \\

\hline
\end{tabular} 
\end{table*}

\begin{figure*}[t]
 \centering
 \mbox
 {
  \hspace{-0.1in}
  \begin{subfigure}[Attention before MAML fast adaptation]{
    \centering
    \includegraphics[width = 0.37 \linewidth]{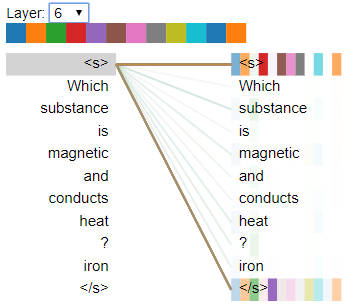}
   }
  \end{subfigure}
  \hspace{0.6in}
  \begin{subfigure}[Attention after MAML fast adaptation]{
    \centering
    \includegraphics[width = 0.37 \linewidth]{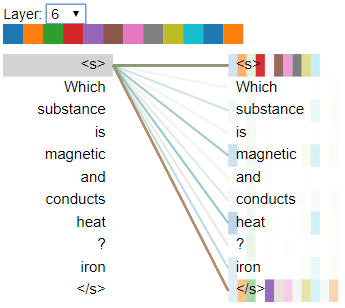}
   }
  \end{subfigure}
 }
 \caption{Attention-head view for few-shot learning model, for the input text \emph{Which substance is magnetic and conducts heat? iron.} The darker color means the higher weights. The output token $<$s$>$ projects to the reasoning result of questions.}
 \label{fig:robust}
\end{figure*}

In order to verify our inference, we conduct a transfer learning experiments for in-depth analysis. We firstly train the classifier on the same training set, then finetune it on visible samples ($k$-shot each class) on test set. \emph{Transfer learning} method in Table 4 shows that under the L1 setting, where the differences between the categories and within categories are large, the accuracy of the MAML-based method achieves 16\% and 12.8\% higher than the transfer method. As the level increases, the numerical difference between the results of the transfer-based method and the MAML-based method decreases, but the MAML-based method still presents obvious advantages. The results prove that meta-classifier can effectively extract meta-features, which ensures excellent generalization performance on different tasks.

\subsection{Model Visualization}

In order to give an insight into the effectiveness of the MAML method on few-shot question classification, we make use of visualization tool \cite{vig2019multiscale} to analyze the attention weights for each word from query question, as shown in Figure 3. 

We can observe that before parameters adaptation, model only attends to the start token and the end token. After quickly fine-tuning the parameters on one supporting sample, the key information (substance, magnetic, heat, iron) of the lengthy question and answer are given higher weights. The terms with the highest attention weights (magnetic and heat) are exactly the most important clues for reasoning label \emph{Properties of Materials}, indicating the MAML method works effectively on few-shot learning problem.

\subsection{Question Answering with Few-shot QC Information}

We incorporate few-shot QC information into reasoning procedure by expanding related QC information on QA input. Figure 4 shows QA performance from L1 to L4, where the \emph{baseline} refers to the model that does not rely on any external information; the \emph{predicted labels and shots} represents the model using predicted information from the few-shot question classification model; the \emph{gold labels and shots} provides the truth label and real relational example questions for test samples; the \emph{top5 corpus} presents the performance of retrieval-based QA method, which relies on the top-5 related sentences from the ARC corpus as background knowledge. By analyzing the curve, we observe that utilizing example questions and labels produced large gains in QA performance, and the QA performance improves as the number of example questions increases.

\begin{figure*}[!h]
\centering
\subfigure[QA performance on L1]{
\begin{minipage}[t]{0.37\linewidth}
\centering
\includegraphics[width = 1\linewidth]{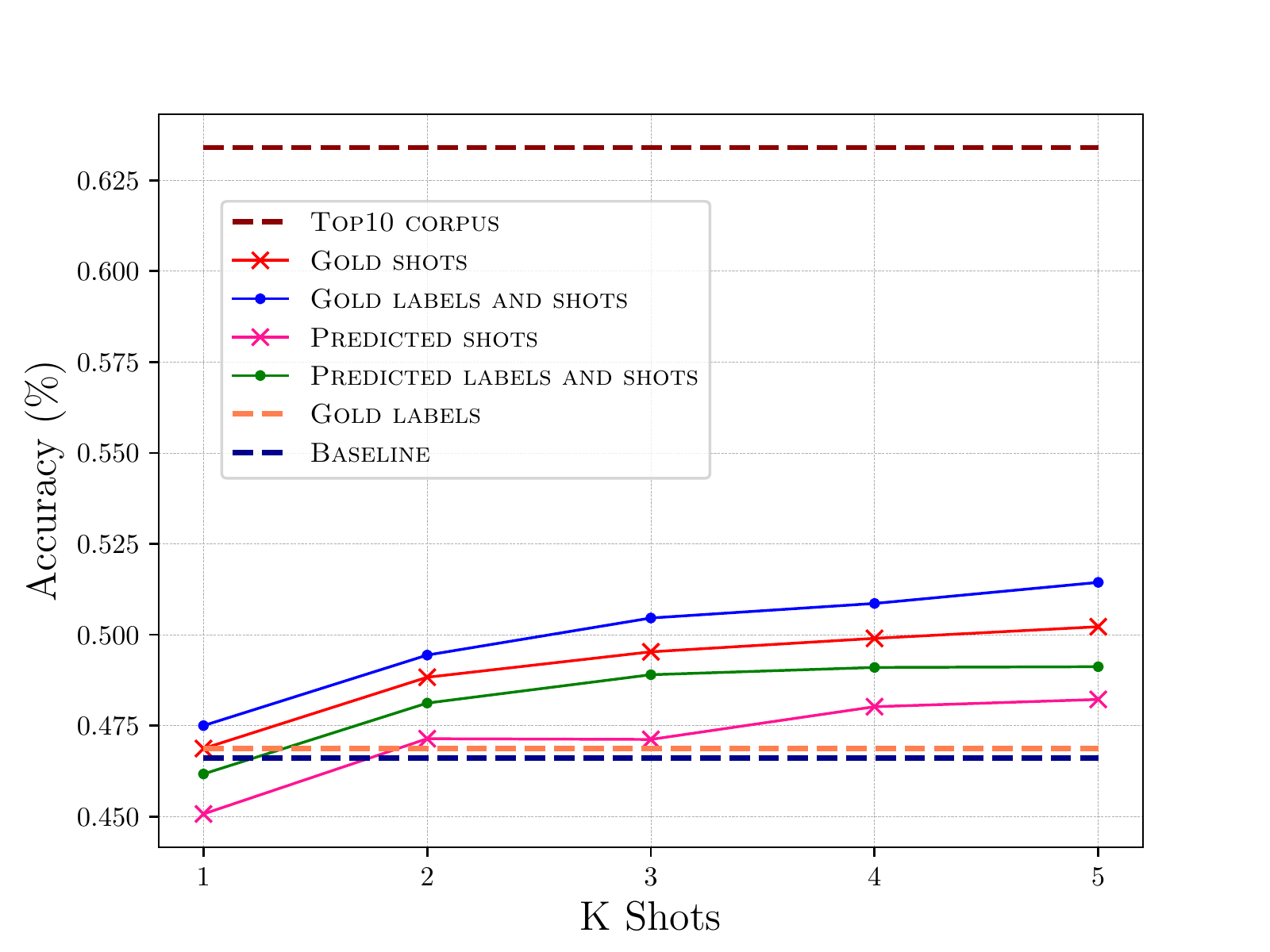}
\end{minipage}%
}%
\hspace{.2in}
\subfigure[QA performance on L2]{
\begin{minipage}[t]{0.37\linewidth}
\centering
\includegraphics[width =  1\linewidth]{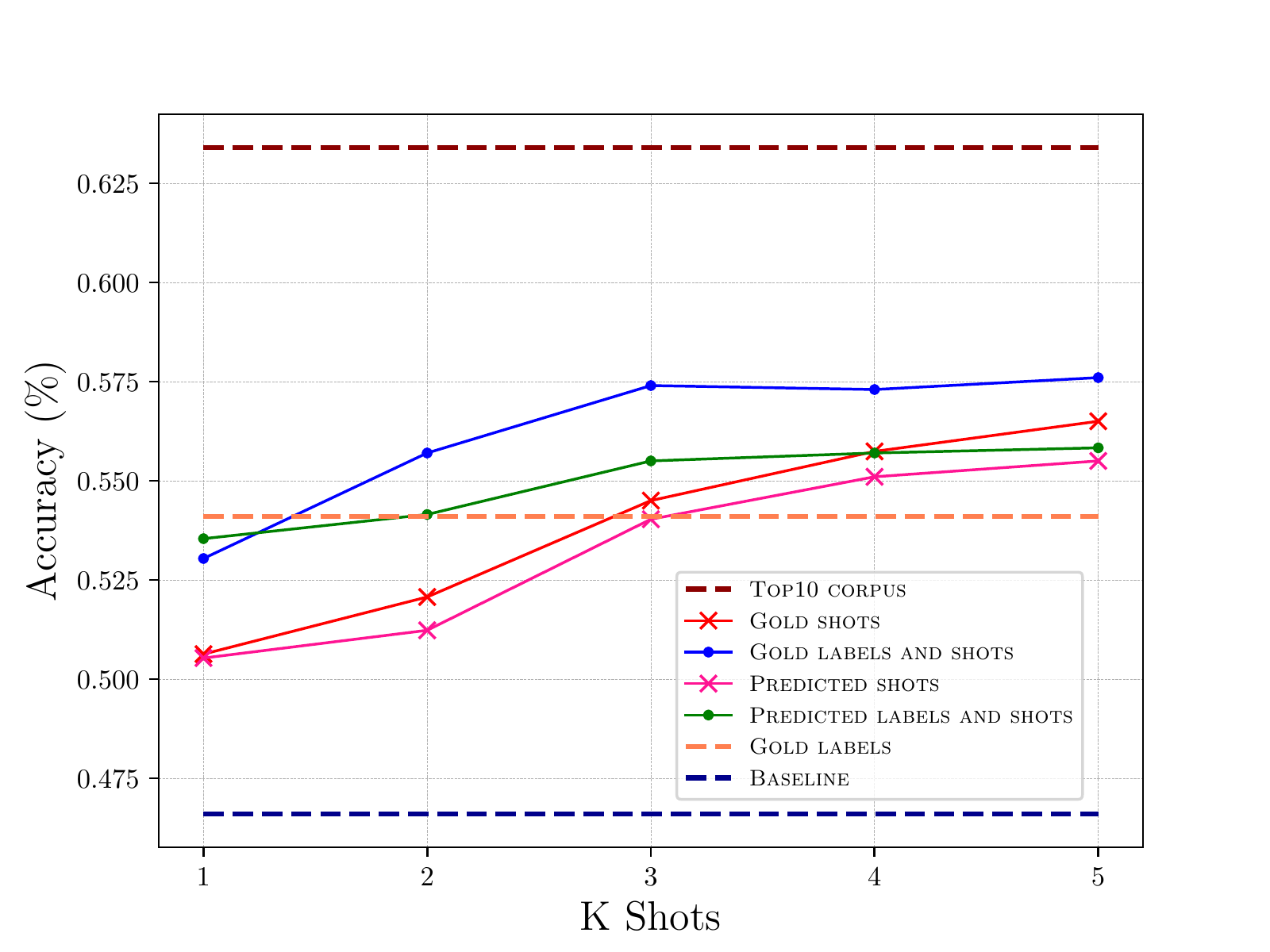}
\end{minipage}%
}%
\hspace{.1in}
\subfigure[QA performance on L3]{
\begin{minipage}[t]{0.37\linewidth}
\centering
\includegraphics[width =  1\linewidth]{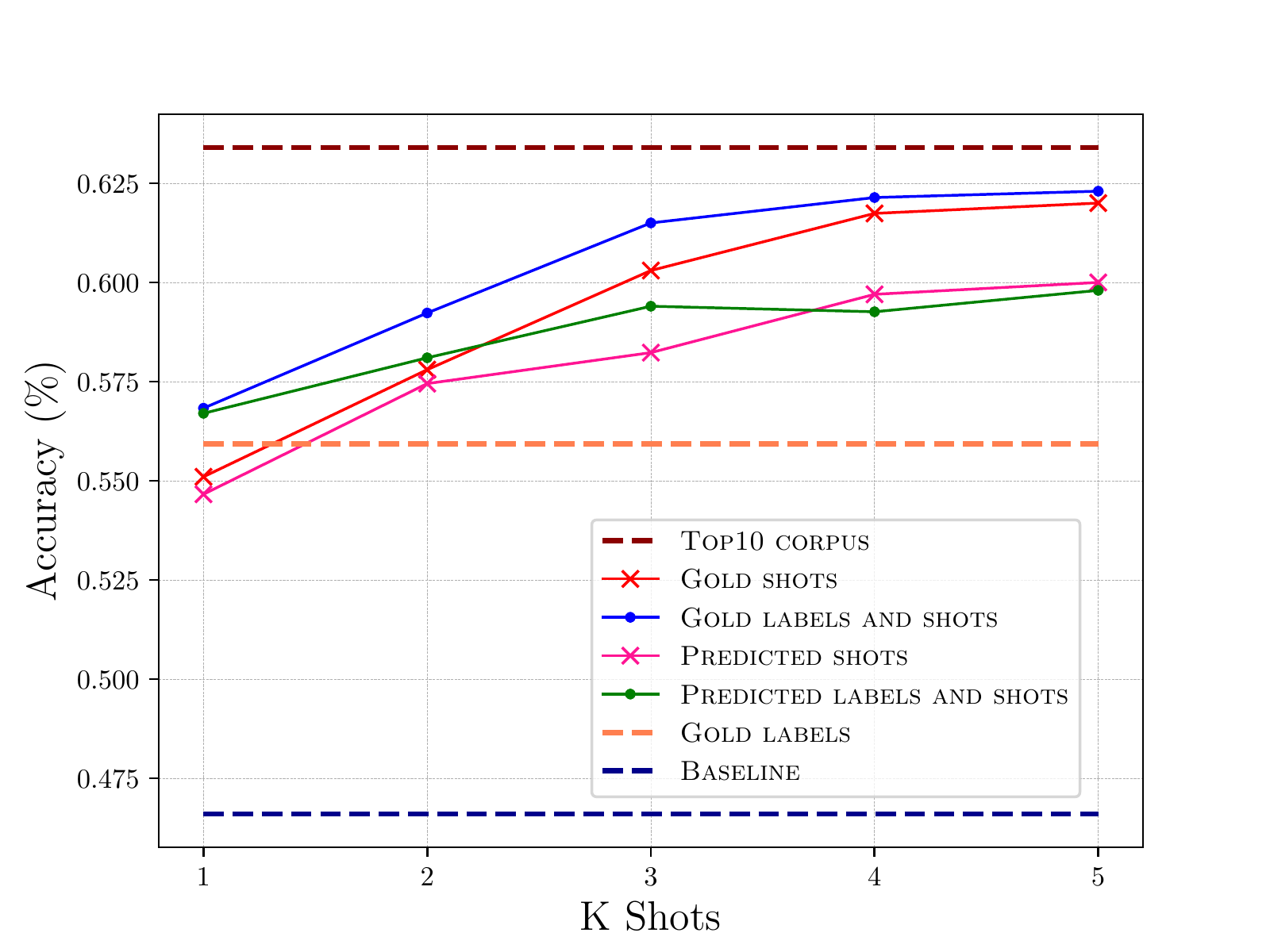}
\end{minipage}%
}%
\hspace{.2in}
\subfigure[QA performance on L4]{
\begin{minipage}[t]{0.37\linewidth}
\centering
\includegraphics[width =  1\linewidth]{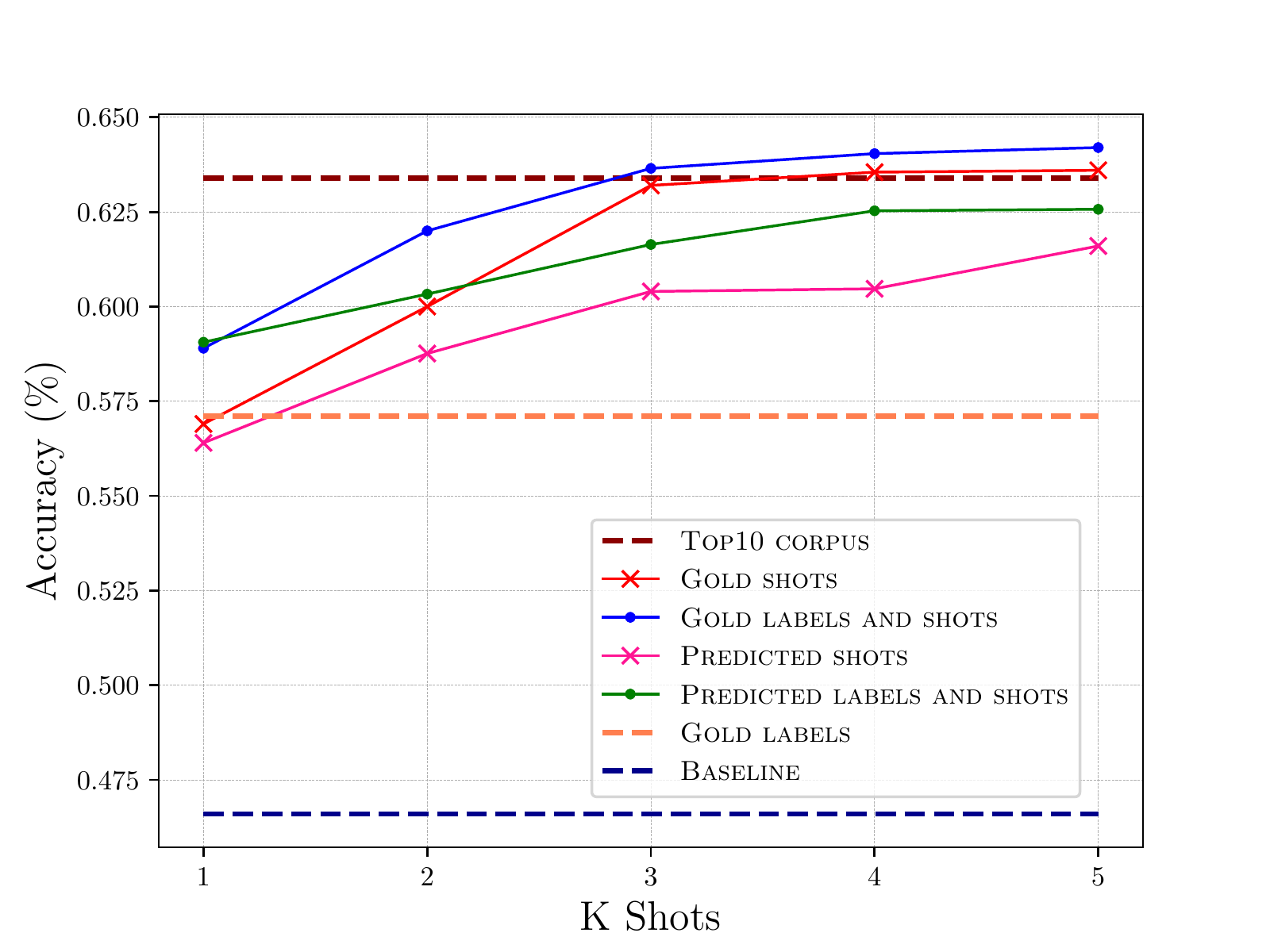}
\end{minipage}
}%
\hspace{-0.1in}
\centering
\caption{Question answering performance based on question classification information from L1 to L4, compared to the baseline model and Top5 retrieval-assisted model.}
\end{figure*}

\begin{table}[!h]
\renewcommand\arraystretch{1.1} 
\renewcommand\tabcolsep{9pt}
\centering  
\caption{QA accuracy with the assistance of 5-shot 5-way question classification information on the ARC science exam, broken down by classification granularity (L1 to L4)}
\begin{tabular}{lcccc}  
\hline
 QC Info & L1 & L2 & L3 & L4 \\ 
 
\hline 
\emph{Baseline} & \multicolumn{4}{c}{$46.6\%$} \\

\hline

Pred Shots & $48.2\%$ &$55.5\%$  & $56.0\%$ &$61.6\%$ \\
\textbf{Pred Labels \& Shots} & $49.1\%$ &$55.8\%$ & $59.8\%$ & $62.6\%$ \\

\hline
Gold Labels & $51.3\%$ &$54.1\%$ & $55.9\%$ & $57.1\%$ \\
Gold Shots & $50.2\%$ &$56.5\%$ & $62.0\%$ & $63.6\%$ \\
\textbf{Gold Labels \& Shots} & $51.4\%$ &$57.6\%$ & $62.3\%$ & $\textbf{64.2\%}$ \\

\hline
\emph{Top5 Corpus} & \multicolumn{4}{c}{$63.4\%$} \\
\hline
\end{tabular} 
\end{table}

From the results of QA performance on L1 to L4, it can be found that the finer the classification information provided by meta-classifier, the more effective information can be obtained by reasoning system. For example, the QA accuracy increases from $49.1\%$ (L1) to $62.6\%$ (L4) with the assistance of predicted labels and $5$ example questions (Table 5). Impressively, with the information given by both gold examples and labels, the MetaQA system reaches $64.2\%$ high accuracy, which outperforms the $63.4\%$ accuracy of retrieval-based method.

From the comparison of \emph{gold} curve and \emph{predicted} curve, a certain proportion of misclassified predictions will lead to incorrect question answering decisions. With the improvement of few-shot QC accuracy from $1$-shot to $5$-shot, the MetaQA system obtains substantial performance gains over predicted information from meta-classifier.

\subsection{Case Study}

Utilizing the example questions and label information can provide an important contextual signal to the reasoning module, which would intelligently orient reasoners to determine the problem domain of the question, and ensure the MetaQA system answering with high confidence and accuracy.

\begin{table}[!h]\small
\renewcommand\arraystretch{1.1}
\renewcommand\tabcolsep{5.0pt}
\centering  
\caption{Case Study}
\begin{tabular}{p{0.12\columnwidth}|p{0.78\columnwidth}}  

\hline 
Test Q & \emph{Female seals usually return to the same beaches year after year to give birth. If they are repeatedly disturbed by humans at those beaches, how will the seals most likely respond?}   \\ 

\\
Ans & (B) They will give birth to more pups. $\times$\\
\hline
\textbf{+Label}  &  Environmental effects on animals \\
Ans   &  (C) They will give birth at different beaches. $\checkmark$ \\

\hline
\textbf{+e.g. Q}   &  \emph{A rainforest area is experiencing a severe drought. As a result, the insect population has decreased. What will the insect-eating birds most likely do? Move to a new area to find food.}\\
Ans   &  (C) They will give birth at different beaches. $\checkmark$  \\

\hline \hline
Test Q & \emph{Fourth graders are planning a roller-skate race. Which surface would be the best for this race?}   \\ 

\\
Ans & (A) gravel $\times$ \\
\hline
\textbf{+Label}  &  Friction \\
Ans   &  (B) sand $\times$ \\

\hline
\textbf{+e.g. Q} & \emph{Which material should be used on a bicycle ramp to increase friction? rough paper}\\
Ans & (C) blacktop. $\checkmark$ \\

\hline
\end{tabular} 
\end{table}

The first example in Table 6 illustrates that the solver chooses the wrong answer (B) based on the question text. Given a label or an example could lead reasoner to the domain of \emph{environmental effects on animals}, and helps to make the right decisions. The second example demonstrates that the solver still selects the wrong answer with the help of label information. However, if the solver is provided by related example questions, it can extract enough information to make inference and finally choose the right answer. From these two examples, we can conclude that our targeted solution - using the label information and same type questions to infer test questions, does improve the question answering performance.

\section{Conclusion}

This paper introduces a new framework MetaQA, which is based on a meta-classifier system and a reasoning system to challenge closed-book science exam. Inspired by cognitive science, two systems complement each other. Meta-classifier adopts meta-learning methods to enable the system to quickly classify new knowledge. The reasoning system uses strong attention mechanism to inference from information given by meta-classifier without suffering a procedure of large corpus retrieval. The experiments show the meta-classifier trained by MAML can be directly used to predict any unseen question types and achieve 85.5\% high classification accuracy. With the help of meta-classifier, MetaQA system achieves gains of up to 17.6\% in ARC science exam, and outperforms the corpus-based approach.

\bibliographystyle{splncs04}
\bibliography{PKAW}

\begin{thebibliography}{10}
\providecommand{\url}[1]{\texttt{#1}}
\providecommand{\urlprefix}{URL }
\providecommand{\doi}[1]{https://doi.org/#1}

\bibitem{clark2015elementary}
Clark, P.: Elementary school science and math tests as a driver for ai: take
  the aristo challenge! In: Twenty-Seventh IAAI Conference (2015)

\bibitem{clark2016my}
Clark, P., Etzioni, O.: My computer is an honor student—but how intelligent
  is it? standardized tests as a measure of ai. AI Magazine  \textbf{37}(1),
  5--12 (2016)

\bibitem{clark2019f}
Clark, P., Etzioni, O., Khot, T., Mishra, B.D., Richardson, K., Sabharwal, A.,
  Schoenick, C., Tafjord, O., Tandon, N., Bhakthavatsalam, S., et~al.:
  From'f'to'a'on the ny regents science exams: An overview of the aristo
  project. arXiv preprint arXiv:1909.01958  (2019)

\bibitem{clark2013study}
Clark, P., Harrison, P., Balasubramanian, N.: A study of the knowledge base
  requirements for passing an elementary science test. In: Proceedings of the
  2013 workshop on Automated knowledge base construction. pp. 37--42. ACM
  (2013)

\bibitem{devlin2018bert}
Devlin, J., Chang, M.W., Lee, K., Toutanova, K.: Bert: Pre-training of deep
  bidirectional transformers for language understanding. arXiv preprint
  arXiv:1810.04805  (2018)

\bibitem{dua2019drop}
Dua, D., Wang, Y., Dasigi, P., Stanovsky, G., Singh, S., Gardner, M.: Drop: A
  reading comprehension benchmark requiring discrete reasoning over paragraphs.
  arXiv preprint arXiv:1903.00161  (2019)

\bibitem{Dual-processing}
Evans, J.S.B.T.: Dual-processing accounts of reasoning, judgment, and social
  cognition. In: Annual Review of Psychology (2008)

\bibitem{finn2017model}
Finn, C., Abbeel, P., Levine, S.: Model-agnostic meta-learning for fast
  adaptation of deep networks. In: Proceedings of the 34th International
  Conference on Machine Learning-Volume 70. pp. 1126--1135. JMLR. org (2017)

\bibitem{godea2018annotating}
Godea, A., Nielsen, R.: Annotating educational questions for student response
  analysis. In: Proceedings of the Eleventh International Conference on
  Language Resources and Evaluation (LREC 2018) (2018)

\bibitem{graves2014neural}
Graves, A., Wayne, G., Danihelka, I.: Neural turing machines. arXiv preprint
  arXiv:1410.5401  (2014)

\bibitem{gu2018meta}
Gu, J., Wang, Y., Chen, Y., Cho, K., Li, V.O.: Meta-learning for low-resource
  neural machine translation. arXiv preprint arXiv:1808.08437  (2018)

\bibitem{hovy2001toward}
Hovy, E., Gerber, L., Hermjakob, U., Lin, C.Y., Ravichandran, D.: Toward
  semantics-based answer pinpointing. In: Proceedings of the first
  international conference on Human language technology research (2001)

\bibitem{jansen2017framing}
Jansen, P., Sharp, R., Surdeanu, M., Clark, P.: Framing qa as building and
  ranking intersentence answer justifications. Computational Linguistics
  \textbf{43}(2),  407--449 (2017)

\bibitem{jansen2018worldtree}
Jansen, P.A., Wainwright, E., Marmorstein, S., Morrison, C.T.: Worldtree: A
  corpus of explanation graphs for elementary science questions supporting
  multi-hop inference. arXiv preprint arXiv:1802.03052  (2018)

\bibitem{khashabi2018question}
Khashabi, D., Khot, T., Sabharwal, A., Roth, D.: Question answering as global
  reasoning over semantic abstractions. In: Thirty-Second AAAI Conference on
  Artificial Intelligence (2018)

\bibitem{lai2017race}
Lai, G., Xie, Q., Liu, H., Yang, Y., Hovy, E.: Race: Large-scale reading
  comprehension dataset from examinations. arXiv preprint arXiv:1704.04683
  (2017)

\bibitem{liu2019roberta}
Liu, Y., Ott, M., Goyal, N., Du, J., Joshi, M., Chen, D., Levy, O., Lewis, M.,
  Zettlemoyer, L., Stoyanov, V.: Roberta: A robustly optimized bert pretraining
  approach. arXiv preprint arXiv:1907.11692  (2019)

\bibitem{minsky1988society}
Minsky, M.: Society of mind. Simon and Schuster (1988)

\bibitem{mishra2017simple}
Mishra, N., Rohaninejad, M., Chen, X., Abbeel, P.: A simple neural attentive
  meta-learner  (2017)

\bibitem{munkhdalai2017meta}
Munkhdalai, T., Yu, H.: Meta networks. In: Proceedings of the 34th
  International Conference on Machine Learning-Volume 70. pp. 2554--2563. JMLR.
  org (2017)

\bibitem{musa2018answering}
Musa, R., Wang, X., Fokoue, A., Mattei, N., Chang, M., Kapanipathi, P., Makni,
  B., Talamadupula, K., Witbrock, M.: Answering science exam questions using
  query reformulation with background knowledge  (2018)

\bibitem{nichol2018reptile}
Nichol, A., Schulman, J.: Reptile: a scalable metalearning algorithm. arXiv
  preprint arXiv:1803.02999  \textbf{2} (2018)

\bibitem{pan2019improving}
Pan, X., Sun, K., Yu, D., Ji, H., Yu, D.: Improving question answering with
  external knowledge. arXiv preprint arXiv:1902.00993  (2019)

\bibitem{peters2018deep}
Peters, M.E., Neumann, M., Iyyer, M., Gardner, M., Clark, C., Lee, K.,
  Zettlemoyer, L.: Deep contextualized word representations. arXiv preprint
  arXiv:1802.05365  (2018)

\bibitem{qiu1993concept}
Qiu, Y., Frei, H.P.: Concept based query expansion. In: Proceedings of the 16th
  annual international ACM SIGIR conference on Research and development in
  information retrieval. pp. 160--169. ACM (1993)

\bibitem{ran2019numnet}
Ran, Q., Lin, Y., Li, P., Zhou, J., Liu, Z.: Numnet: Machine reading
  comprehension with numerical reasoning. arXiv preprint arXiv:1910.06701
  (2019)

\bibitem{roberts2014automatically}
Roberts, K., et~al., K.: Automatically classifying question types for consumer
  health questions. In: AMIA Annual Symposium Proceedings. vol.~2014, p.~1018.
  American Medical Informatics Association (2014)

\bibitem{santoro2016one}
Santoro, A., Bartunov, S., Botvinick, M., Wierstra, D., Lillicrap, T.: One-shot
  learning with memory-augmented neural networks. arXiv preprint
  arXiv:1605.06065  (2016)

\bibitem{empirical}
Sloman, S.A.: The empirical case for two systems of reasoning. psychological
  bulletin (1996)

\bibitem{vig2019multiscale}
Vig, J.: A multiscale visualization of attention in the transformer model.
  arXiv preprint arXiv:1906.05714  (2019)

\bibitem{xu2019multi}
Xu, D., et~al., J.: Multi-class hierarchical question classification for
  multiple choice science exams. arXiv preprint arXiv:1908.05441  (2019)

\end{thebibliography}

\end{document}